\documentclass[pdflatex,sn-mathphys-ay, iicol]{sn-jnl}
\usepackage{multirow}%
\usepackage{amsmath,amssymb,amsfonts}%
\usepackage{amsthm}%
\usepackage{mathrsfs}%
\usepackage[title]{appendix}%
\usepackage{xcolor}%
\usepackage{textcomp}%
\usepackage{manyfoot}%
\usepackage{booktabs}%
\usepackage{algorithm}%
\usepackage{algorithmicx}%
\usepackage{algpseudocode}%
\usepackage{listings}%
\usepackage{bm}
\usepackage{makecell}

\theoremstyle{thmstyleone}%

\theoremstyle{thmstyletwo}%
\theoremstyle{thmstylethree}%
\begin{document}

\title[Article Title]{SimTO: A \textcolor{black}{two-stage,} simulation-\textcolor{black}{driven} topology optimization framework for bespoke soft robotic grippers}

\author*[1,2]{\fnm{Kurt} \sur{Enkera}}\email{kurt.enkera@csiro.au}
\author[1]{\fnm{Josh} \sur{Pinskier}}\email{josh.pinskier@csiro.au}
\author[2]{\fnm{Marcus} \sur{Gallagher}}\email{marcusg@uq.edu.au}
\author[1]{\fnm{David} \sur{Howard}}\email{david.howard@csiro.au}

\affil[1]{\orgdiv{CSIRO Robotics}, \orgname{CSIRO}, \orgaddress{\street{1 Technology Ct}, \city{Brisbane}, \postcode{4069}, \state{Queensland}, \country{Australia}}}
\affil[2]{\orgdiv{School of Electrical Engineering and Computer Science}, \orgname{The University of Queensland}, \orgaddress{\street{Sir Fred Schonell Drive}, \city{Brisbane}, \postcode{4072}, \state{Queensland}, \country{Australia}}}

\abstract{Soft robotic grippers are essential for grasping delicate, geometrically complex objects in manufacturing, healthcare and agriculture. However, existing designs struggle to grasp \emph{feature-rich} objects with high topological variability, including gears with sharp tooth profiles on automotive assembly lines, corals with fragile protrusions, or vegetables with irregular branching structures like broccoli. Unlike simple geometric primitives such as cubes or spheres, feature-rich objects lack a clear ``optimal'' contact surface, making them both difficult to grasp and susceptible to damage. Safe handling of such objects therefore requires \emph{specialized} soft grippers whose morphology is tailored to the object's features. Topology optimization offers a promising approach for producing specialized grippers, but its utility is limited by the need for pre-defined load cases. For soft grippers, these loads arise from hundreds of unpredictable gripper-object contact forces during grasping and are unknown \emph{a priori}. To address this problem, we introduce \textbf{SimTO}, a two-stage, \textbf{sim}ulation-driven \textbf{T}opology \textbf{O}ptimization framework that automatically extracts load cases from a dynamic, contact-rich grasping simulation before performing classical topology optimization, eliminating the need for manual load specification. Given an arbitrary feature-rich object, \emph{SimTO} produces highly customized soft grippers with fine-grained morphological features tailored to the object geometry. \textcolor{black}{Physical experiments confirm that our specialized grippers achieve higher grasp forces than a generalist design produced by conventional topology optimization methods, while numerical experiments show that they achieve high grasp success rates across varying object poses and strong generalization to a set of unseen objects.}}


\maketitle


\section{Introduction}\label{sec1}

\begin{figure*}[th!]
    \centering
    \includegraphics[width=0.92\textwidth]{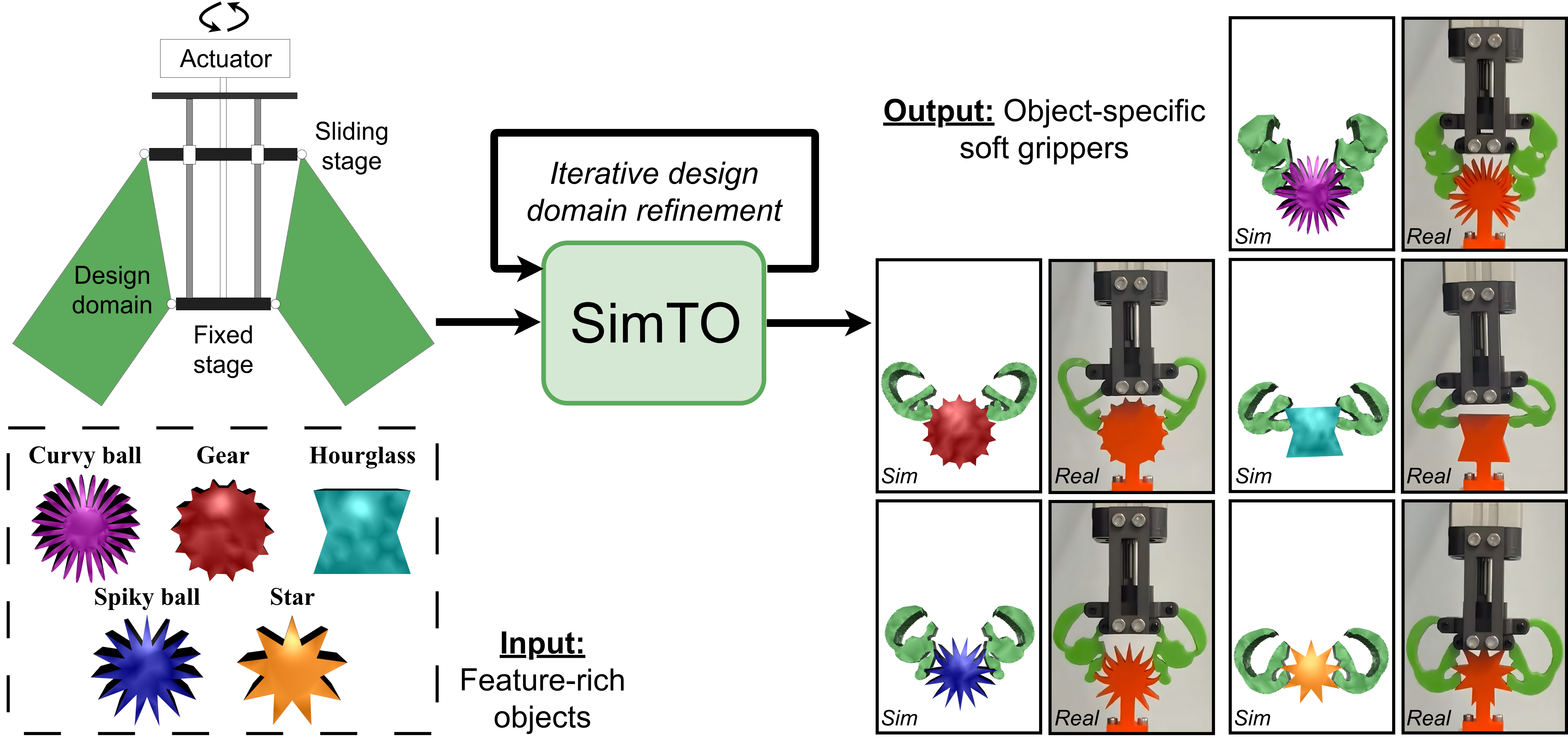}
    \caption{\emph{The SimTO framework.} \textbf{Left:} Inputs to SimTO include (i) a deformable, feature-rich object and (ii) a soft gripper whose dynamic grasping behaviour can be simulated. In this work, we used a soft gripper design scheme inspired by \citet{liu2018_TObenchmark_v2}, whose end-effectors are soft fingers actuated by the compression of a sliding stage. \textbf{Right:} Given an arbitrary feature-rich object, SimTO generates bespoke soft fingers which conform to that object's shape. \textcolor{black}{The resulting grippers exhibit significantly higher peak grasp forces (Sec.~\ref{expval}) than the generalist design by \citet{liu2018_TObenchmark_v2}.}} 
    \label{fig:fig1}
\end{figure*}

Soft robotic grippers are typically designed as \emph{generalists} intended to handle a variety of objects \citep{Brown2010, pneunet2011, Pfaff2011FinRay, neuralphys2025}, while \emph{specialist} designs tailored to a constrained set of objects are underexplored \citep{machines12120887}. However, most industrial soft grasping applications - including fruit harvesting or repetitive pick-and-place operations in warehouses - target only a constrained set of objects \citep{machines12120887}, highlighting the need for specialist designs in industry. Furthermore, generalist designs often struggle to grasp \emph{feature-rich} objects with high topological variability, including gears on automotive assembly lines, corals with fragile protrusions, and vegetables with branching structures like cauliflower and broccoli. These limitations highlight the need for a design framework capable of generating specialist soft grippers for feature-rich objects. 

Designing specialized soft grippers is challenging because it requires detailed knowledge of the contact interactions between the gripper and object \citep{Birglen2008UnderactuatedHands} - data that is often difficult or infeasible to obtain. Topology optimization (TO) is a powerful structural design framework and a strong candidate for producing specialized soft grippers \citep{BendsoeSigmund2004}, but its applicability is limited by the need for pre-defined load cases. For soft grippers, these loads arise from hundreds of unpredictable gripper-object contact forces during grasping and are unknown \emph{a priori}. Recently, parametric optimization approaches have emerged which accurately model rich, time-varying gripper-object contact forces \citep{Gjoka2024, neuralphys2025}. However, few works incorporate these contact forces into their optimization objective, and many use simple design parameterizations that hinder the generation of designs with diverse features. 

\textcolor{black}{Outside of optimization-based methods, a simple way to generate object-specific grippers is via `heuristic imprinting' \citep{fit2form}, where the target object's shape is cut from a rectangular block to produce a gripper that closely matches the object geometry. While highly specialized, imprint designs are extremely sensitive to variations in object pose and therefore lack robustness in practice.}

\textcolor{black}{In response, we introduce \textbf{SimTO}, a two-stage, \textbf{sim}ulation-driven \textbf{T}opology \textbf{O}ptimization framework that first performs (1) dynamic soft grasping simulations in a standalone robotics simulator. Rich load cases are then automatically extracted from these simulations and used for (2) classical topology optimization.} This approach is particularly advantageous for problems where load cases are complex or difficult to obtain. Our \textbf{key contributions} are summarized as follows: 

\begin{itemize}
    \item We present \emph{SimTO}, a computational design algorithm that removes the need for pre-defined load cases in classical topology optimization by automatically extracting them from dynamic physics simulations instead.
    
    \item For problems involving design-dependent loads, \emph{SimTO} enables an iterative optimization process that continually refines both the design and its corresponding loads. By repeatedly alternating between dynamic, simulation-based load extraction and topology optimization, we incrementally arrive at highly specialized soft grippers for feature-rich objects.
    
    \item \textcolor{black}{Experiments show that our specialized grippers achieve high grasp success rates across varying object poses, strong generalization to a set of unseen objects and higher grasp forces than a generalist design produced by conventional TO methods.}
\end{itemize}

\section{Literature Review}\label{sec2}

Computational design frameworks for soft grippers share four fundamental components: (i) a design representation, (ii) a simulation method for performance evaluation, (iii) an objective function, and (iv) an optimization algorithm \citep{PINSKIER2024303}. A design representation is defined by an $N$-dimensional vector of design variables, $\boldsymbol{\rho}$, that characterizes a soft gripper's morphology, while the set of all possible vectors $\boldsymbol{\rho}$ forms an $N$-dimensional \emph{design space}. Design performance is evaluated in simulation and quantified by an objective function, $f(\boldsymbol{\rho})$, which is then extremized by either a gradient-based or gradient-free optimization algorithm. While gradient-based methods are efficient at finding local optima in high-dimensional design spaces, gradient-free methods offer fewer convergence guarantees \citep{PINSKIER2024303}.

Both topology optimization and parametric optimization approaches can be used for soft gripper design, with each offering different strengths and weaknesses. Topology optimization is a powerful, gradient-based optimization framework that can obtain locally optimal solutions within high-dimensional design spaces \citep{BendsoeSigmund2004}, making it a strong candidate for producing specialized soft grippers. Conversely, many parametric design approaches \citep{pneu_soft2022, Navarro2023, YaoSPADA2024, Navez2024} use gradient-free evolutionary algorithms to optimize only a limited set of design variables, preventing the emergence of specialized designs with fine-grained features.

Within topology optimization, however, a critical limitation exists due to the finite element formulation that is typically used. In SIMP-based topology optimization (Solid Isotropic Material with Penalization), designs are represented as a 2D or 3D grid of finite elements \citep{BendsoeSigmund2004}. Each element $e$ is assigned a design variable $\rho_e \in [0,1]$, with $\rho_e = 1$ indicating solid material and $\rho_e = 0$ representing the absence of material. Design performance is evaluated using the finite element method, which crucially requires the pre-specification of boundary conditions and applied loads. In many applications, \textbf{these loads cannot be easily determined}. For example, when optimizing the internal structure of an aircraft wing, \citet{Aage2017} required rich load cases obtained from expensive NASA wind-tunnel experiments. Similarly, when designing soft grippers specialized for feature-rich objects, the true load cases are not known \emph{a priori}.

To our knowledge, all soft grippers designed via topology optimization \citep{Mankame2004, liu2018_TObenchmark_v2, caasenbrood2020, cableTO2020, liu2023, Huang2023ClampingForce, pinskier2024} have largely overlooked the \emph{time-dependent} contact interactions that occur between the gripper and object during grasping. Instead, most works have used generic, manually prescribed `dummy' loads \citep{frecker1999} that define fixed trajectories for grippers to deform along. To date, this generic dummy load approach has only produced generalist designs that lack the specialized morphological features needed to securely grasp feature-rich objects. \textcolor{black}{Although \citet{Mankame2004} and \citet{Huang2023ClampingForce} optimized soft grippers while explicitly modelling gripper-object contact, both works only considered \emph{static} contact scenarios rather than rich, dynamic contact.} 

To address these gaps, we introduce \emph{SimTO}, a framework that replaces generic, manually prescribed dummy loads with rich gripper-object contact forces automatically extracted from dynamic soft grasping simulations. 

\begin{figure*}[th!]
    \centering
    \includegraphics[width=0.8\textwidth]{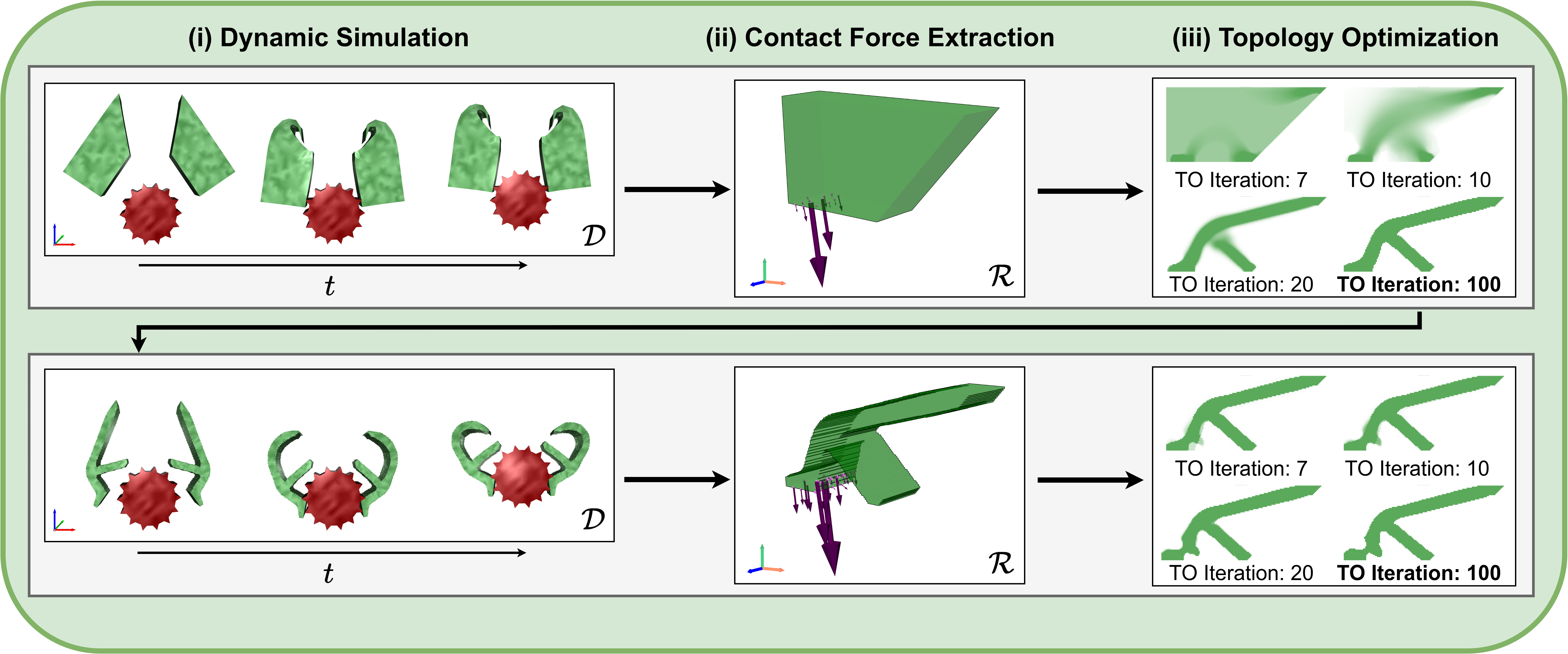}
    \caption{\textit{The SimTO method.} This figure shows two iterations of SimTO. Each iteration consists of: (i) a dynamic simulation of a 3D compliant mechanism in the deformed frame $\mathcal{D}$; (ii) contact force extraction, where the simulated 3D forces are rotated into $\mathcal{R}$ and projected onto the $x\!-\!y$ plane; and (iii) topology optimization of the undeformed 2D design using these in-plane forces. The resulting 2D design is then extruded to initialize the next iteration.}
    \label{fig:method}
\end{figure*}

\section{Method}\label{method}

SimTO is a design optimization algorithm for compliant mechanisms - flexible monolithic structures that achieve motion through elastic deformation under loading \citep{frecker1999}. The framework is designed for mechanisms that experience time-varying, design-dependent contact forces while performing a specified task (e.g., grasping an object in the case of a gripper). Each SimTO iteration consists of three stages: (i) dynamic simulation of the structure's rich contact behaviour, (ii) contact force extraction, and (iii) topology optimization using a contact-aware objective function (Fig.~\ref{fig:method}). 

As the designs and hence simulated contact forces evolve with each solve, performing SimTO over multiple iterations yields increasingly specialized mechanisms. As shown in Fig.~\ref{fig:method}, simulating an initial design produces a set of contact forces which automatically define task-specific deformation directions for topology optimization. Simulating the updated design - whose morphology is now better adapted to the task - produces \emph{new} contact interactions and, consequently, \emph{new} deformation directions, leading to further morphological refinement. In the remainder of this section, we explain stages (i) - (iii) of SimTO in further detail.
    
\subsection{Dynamic simulation}

We used Taccel \citep{taccel2025} to perform dynamic soft grasping simulations due to its ability to accurately model time-dependent deformation and frictional contact among multiple interacting Neo-Hookean solids. Each simulation was initialized with the following gripper parameters: material stiffness $E_g$, Poisson's ratio $\nu_g$, \textcolor{black}{friction coefficient $\mu_g$, mass density $\varrho_g$ ($\text{kg}/\text{m}^3$),} a tetrahedral mesh $\mathcal{M}_g$ of the left soft finger with $N_g$ nodes (later duplicated to form the right soft finger), and the initial gripper grasp pose $\mathcal{P}_g$. Analogous quantities were also defined for an arbitrary feature-rich object: $E_o$, $\nu_o$, \textcolor{black}{$\mu_o$, $\varrho_o$,} $\mathcal{P}_o$ and $\mathcal{M}_o$. 

In each dynamic grasping simulation, the sliding stage of a soft robotic gripper (Fig.~\ref{fig:fig1}) was first compressed by $d_c$ millimetres, and then the gripper was lifted by $d_l$ millimetres. Each simulation lasted $t$ seconds and consisted of $N_t$ timesteps. During the first $t/2$ seconds, gripper compression was achieved by fixing the displacements of nodes on the fixed port of the design domain (Fig.~\ref{fig:designdomain}) to $0$, while input port nodes were compressed by $2 \,d_c/N_t$ millimetres per timestep, resulting in a smooth compression of the sliding stage by $d_c$ millimetres. In the final $t/2$ seconds, both input and output port nodes were lifted in the $z$-direction by $2 \,d_l/N_t$ millimetres per timestep, corresponding to a smooth lifting of the entire gripper by $d_l$ millimetres. In all examples presented in this paper, we set $d_c = 80\,\text{mm}$, $\nu_g = \nu_o = 0.4$, \textcolor{black}{$\varrho_g = \varrho_o = 1000 \, \text{kg}/\text{m}^3$, $\mu_g = 1.0$ and $\mu_o = 0.3$.}

\subsection{Contact force extraction}\label{sec:contact}

Dynamic simulations were performed in a deformed frame $\mathcal{D}$, while classical topology optimization was carried out in a 2D subspace of an undeformed frame $\mathcal{R}$ (Fig.~\ref{fig:method}). \textcolor{black}{In dynamic simulations, contact forces are time-varying and multiple forces with different directions can act at a given spatial location, yielding a large collection of 3D forces defined over both space and time. In contrast, classical TO only allows a single load vector to be assigned to each node in the discretized design domain. Consequently, to perform classical TO using simulated, time-varying contact forces, it is necessary to reduce this collection to a set $F$ containing at most one single 2D contact force per node. We hypothesize that such a reduced set $F$ exists which sufficiently characterizes the grasp for topology optimization.}

To obtain $F$ from a given dynamic simulation, we first extracted a set of $\widetilde{N}_f$ contact forces, $\widetilde{F} =\{\widetilde{\bm{f}}_1,\ldots, \widetilde{\bm{f}}_{\widetilde{N}_f}\}$, where each $\widetilde{\bm{f}}_i \in \mathbb{R}^3$ is a contact force vector expressed in $\mathcal{R}$. From this set, we then constructed a reduced set of $N_f \leq \widetilde{N}_f$ two-dimensional forces, $F = \{\bm{f}_1,\ldots, \bm{f}_{N_f}\}$, with $\bm{f}_i \in \mathbb{R}^2$ (Fig.~\ref{fig:designdomain}). To explain this process in detail, it is useful to consider a single dynamic simulation, timestep by timestep. 

At each timestep, contact forces were filtered to ensure they contributed to a secure grasp. Specifically, we only retained forces that (i) applied a positive normal force on the object and (ii) resisted slip. To implement this scheme, after each timestep we used the Kabsch algorithm \citep{kabsch1976} to approximate a rotation matrix that mapped 3D contact forces from $\mathcal{D}$ to $\mathcal{R}$. We only kept forces with a negative $y$-component and a positive $x$-component in $\mathcal{R}$, corresponding to forces satisfying conditions (i) and (ii). To prevent our TO objective function from being dominated by transient or weak contact forces, \textcolor{black}{we used the following update rule}: if a force was already recorded for a specific gripper node, it was replaced by a newly detected force satisfying (i) and (ii) only if the new force had a \textcolor{black}{larger $x$-component}. This procedure yielded the set of $\widetilde{N}_f$ gripper-object contact forces $\{\widetilde{\bm{f}}_1,\ldots, \widetilde{\bm{f}}_{\widetilde{N}_f}\}$.

To obtain the reduced set $F$ from $\widetilde{F}$, we first rounded the $x$-, $y$-, and $z$-coordinates of all $\widetilde{N}_f$ contact points to the nearest millimetre. Because topology optimization was performed in the $x\!-\!y$ plane of $\mathcal{R}$, we grouped all 3D contact forces sharing a common $(x,y)$ coordinate (i.e., forces lying along the same line of constant $z$). For each group, we averaged the 3D force vectors and subsequently set the $z$-component of the resulting vector to zero, yielding a 2D force vector $\bm{f}$ located at $(x,y)$.

\subsection{Topology optimization of a contact-aware objective function}

\begin{figure}[bh!]
    \centering
    \includegraphics[width=0.45\textwidth]{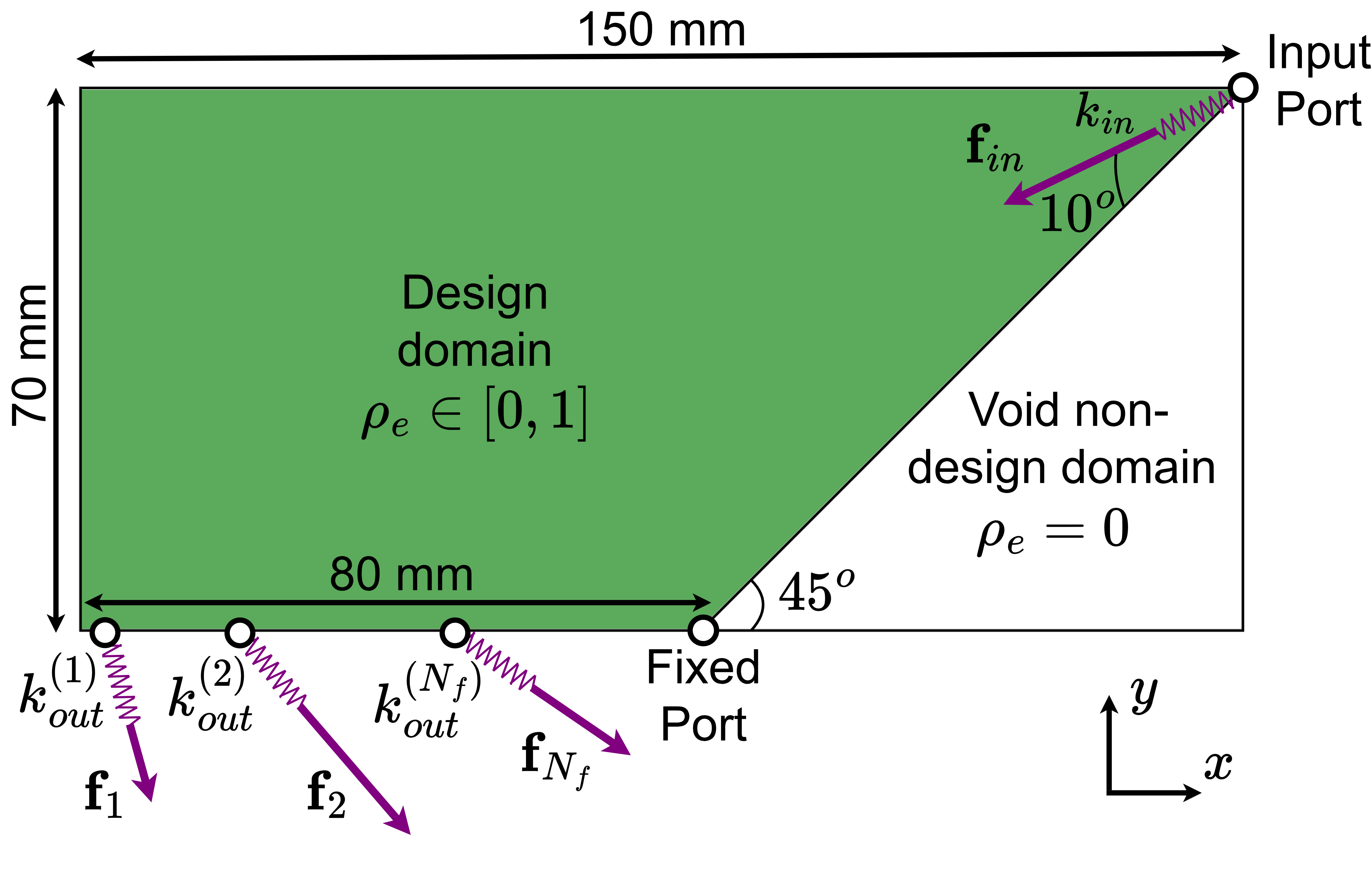}
    \caption{Our design domain is based on that of \citet{liu2018_TObenchmark_v2}, but with three key differences. First, whereas they manually prescribed two dummy loads that encouraged designs to deform solely along the negative $y$-axis, ours deform along $N_f$ simulated gripper-object contact force directions. Second, we rotated the input force $\mathbf{f}_{in}$ by an additional $10^{\circ}$. Third, we did not enforce material retention along the lower $80\mathrm{mm}$ edge of the design domain.}
    \label{fig:designdomain}
\end{figure}

Figure~\ref{fig:designdomain} shows the design domain, boundary conditions and simulated loads used to optimize a single soft finger. We used a conventional objective function for compliant mechanisms that simultaneously encourages designs with sufficient stiffness and deformability in user-specified directions \citep{frecker1999}. However, whereas prior works manually specify a small number of generic deformation directions, we instead used our simulated contact forces $\{\bm{f}_1,\ldots, \bm{f}_{N_f}\}$ to automatically define \emph{object-specific} deformation directions. 

\textcolor{black}{Unlike most similar works \citep{liu2018_TObenchmark_v2, liu2023, Huang2023ClampingForce}, we did not enforce a fixed contact surface. Instead, the contact surface geometry was free to evolve throughout each SimTO iteration. Void regions of material were incapable of generating contact forces during dynamic simulations, but could re-acquire material in later rounds of topology optimization if required.}

For each $\bm{f}_i \in \{\bm{f}_1,\ldots, \bm{f}_{N_f}\}$, we maximized the mutual potential energy (MPE) associated with load $i$ whilst minimizing its corresponding strain energy (SE) \citep{frecker1999}. Maximizing the MPE for load $i$ encourages deformation in the direction of $\bm{f}_i$ at the associated $(x,y)$ contact point, whereas minimizing SE provides adequate stiffness under the actuating input load \citep{frecker1999}. In our work, the $150 \,\text{mm} \times 70\,\text{mm}$ design domain from Fig.~\ref{fig:designdomain} was discretized into $150 \times 70 = 10,500$ square finite elements and a density-based TO approach was used \citep{BendsoeSigmund2004}. Each element $e$ was assigned a material density $\rho_e$, and its Young's modulus was given by:
\begin{equation}
\label{eq:constitutivelaw}
        E(\rho_e) = E_{\text{min}} + \rho_{e}^{p}(E_0 - E_{\text{min}}), 
\end{equation}
where $\rho_e \in [0,1]$, $E_0$ is the normalized material stiffness, $E_{\text{min}} = 10^{-9}$ and $p = 3$. The global stiffness matrix $\mathbf{K}(\boldsymbol{\rho})$ was assembled from the element stiffness matrices, $\mathbf{K}_e^{0}$, and a global stiffness matrix for numerical springs, $\mathbf{K}_s$, via:
\begin{equation}
\label{eq:stiffnessmatrix}
\mathbf{K}(\boldsymbol{\rho}) = \mathbf{K}_s + \sum_{e} E(\rho_e)\mathbf{K}_e^{0}
\end{equation}
where the matrix $\mathbf{K}_s$ introduces standard numerical springs at the input and output ports of the design domain, with spring constants $k_\text{out}^{(i)} = 0.1$ for $i = 1,\ldots,N_f$ and $k_\text{in} = 0.1$. The full optimization problem was formulated as:
\begin{equation*}
\label{eq:optimization}
\begin{aligned}
\operatorname*{argmax}_{\boldsymbol{\rho}} \quad f(\boldsymbol{\rho})
&= \sum_{i=1}^{N_f} \lVert\bm{f}_i\rVert \,\frac{\text{MPE}_i}{\text{SE}_i} \\
&= \sum_{i=1}^{N_f} \lVert\bm{f}_i\rVert
   \,\frac{\mathbf{U}_i^{T}\mathbf{K}(\boldsymbol{\rho})\mathbf{U}_{\text{in}}}
        {\mathbf{U}_i^{T}\mathbf{K}(\boldsymbol{\rho})\mathbf{U}_i} \\[1pt]
\end{aligned}
\end{equation*}
\begin{equation}
\label{eq:optimization}
\begin{aligned}
\text{subject to} \quad
&\frac{V(\boldsymbol{\rho})}{V_0} = v_f, \\
&\mathbf{K}(\boldsymbol{\rho})\,\mathbf{U}_{\text{in}} = \mathbf{F}_{\text{in}}, \\
&\mathbf{K}(\boldsymbol{\rho})\,\mathbf{U}_i = \mathbf{F}_i, 
\quad i = 1,\ldots,N_f.
\end{aligned}
\end{equation}
Here, $\bm{f}_i$ refers to the $i$th simulated contact force in $\{\bm{f}_1,\ldots, \bm{f}_{N_f}\}$, $\mathbf{F}_i$ is the corresponding global force vector containing only this contact force and zeros elsewhere, and $\mathbf{U}_i$ is the global displacement vector. Furthermore, $\mathbf{U}_{\text{in}}$ is the global displacement field produced by the input force $\mathbf{F}_{\text{in}}$, derived from the unit-magnitude force $\bm{f}_{\text{in}}$ shown in Fig.~\ref{fig:designdomain}. \textcolor{black}{The input force, $\mathbf{F}_{\text{in}}$, and $N_f$ output forces are associated with separate equilibrium equations, following common practice in compliant mechanism design \citep{Sigmund1997, frecker1999, liu2018_TObenchmark_v2}.} Finally, the terms $V(\boldsymbol{\rho})$ and $V_0$ represent the material volume and total design domain volume, while $v_f$ is the target volume fraction.

\begin{figure*}[th!]
    \centering  \includegraphics[width=0.88\textwidth]{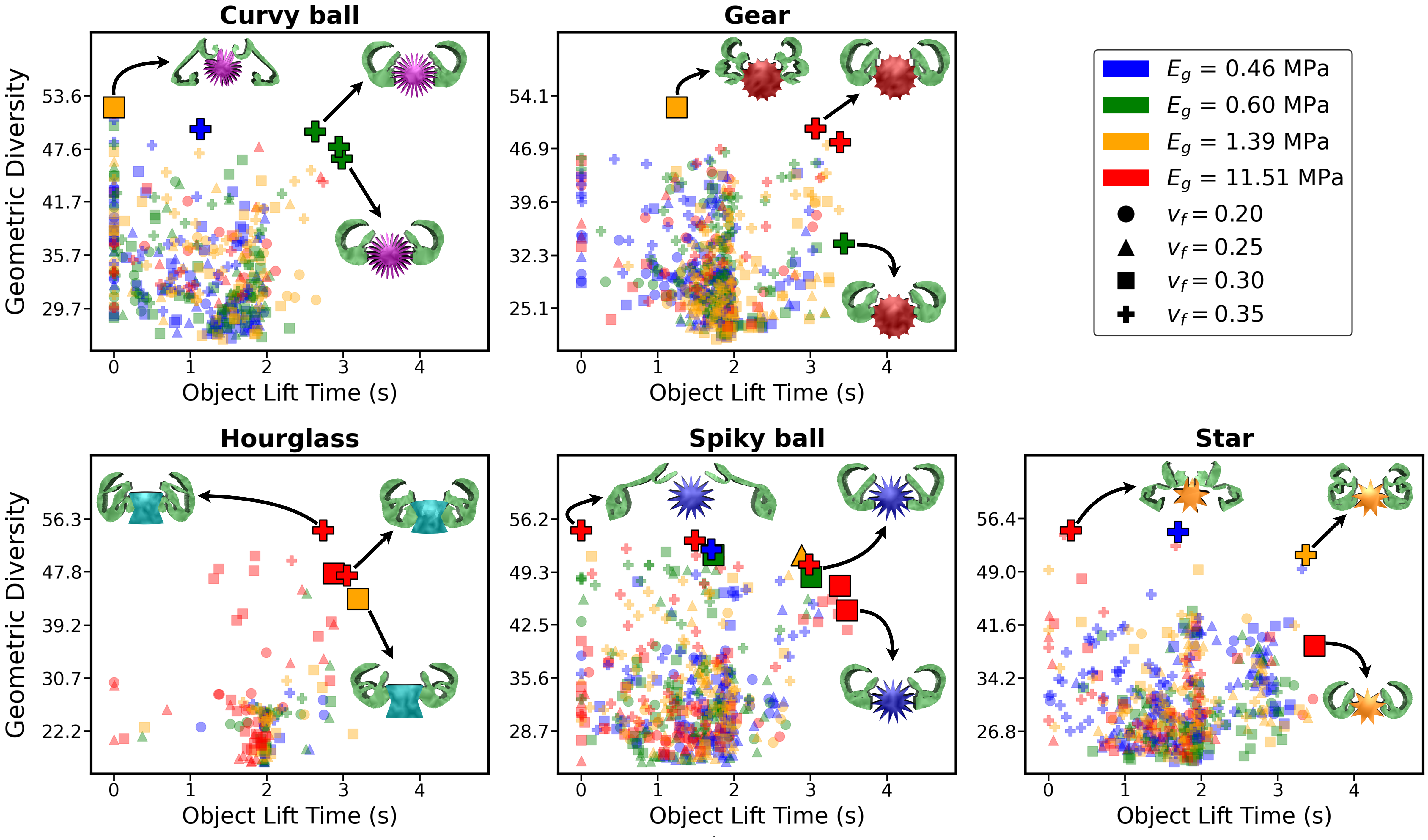}
    \caption{Numerical results of SimTO optimization runs. Each plot compares the geometric diversity and object lift time of all grippers. Designs on the resulting Pareto front are indicated by enlarged markers with black outlines.}
    \label{fig:results}
\end{figure*}

We used a density filter to mitigate checkerboard patterns in optimized designs and encourage mesh-independent solutions \citep{88linecode2011}, while the optimality criteria method \citep{BendsoeSigmund2004} was used to solve (\ref{eq:optimization}). This framework was implemented in a Python-based optimization loop similar to that developed by \citet{88linecode2011}. In this work, topology optimization was terminated when either the change in the material-density design variables fell below one percent or after 100 iterations. Because a single SimTO run may involve multiple rounds of topology optimization, this 100-iteration limit was imposed for computational efficiency. In most cases, this limit was sufficient to achieve high-performing designs.

\section{Numerical Results}\label{results}

\subsection{Generating object-specific soft grippers}

Using SimTO, we generated 2071 bespoke soft grippers (Fig.~\ref{fig:results}) for five different feature-rich objects: a \emph{curvy ball}, \emph{gear}, \emph{hourglass}, \emph{spiky ball} and \emph{star} (Fig.~\ref{fig:fig1}). For each object, we performed a parameter sweep across four volume fractions, $v_f \in \{0.20,0.25,0.30, 0.35\}$, and four 3D-printable material moduli for both the gripper and object, $E_g, E_o \in \{0.46,0.60,1.39, 11.51\} \,\text{MPa}$, resulting in 64 optimization runs per object. As variations in $E_g$, $E_o$ and $v_f$ can significantly impact the simulated gripper-object contact interactions - with $E_g$ and $E_o$ controlling the relative stiffness between the gripper and object, and $v_f$ determining the gripper's total volume and geometry - we expected different optimization runs to yield designs with unique morphological features.

Each run concluded after 20 SimTO iterations or when two successive designs, $\bm{\rho}^i$ and $\bm{\rho}^{i+1}$, satisfied the convergence criterion $\sum_e(\rho_e^{i+1} - \rho_e^{i})^4 < \epsilon$. We set $\epsilon = 10$, an empirically determined value that ended the optimization once two consecutive SimTO iterations produced sufficiently similar designs. However, because the design-dependent nature of the optimization creates a highly non-linear process, designs generated within a given SimTO run were not guaranteed to \emph{monotonically} improve grasping performance. This meant `converged' designs did not always possess the most effective features for specialized grasping. To capture the full range of high-performing designs, we analyzed the performance of all designs generated across every SimTO iteration (Fig.~\ref{fig:results}). 

\newpage

While the theoretical maximum number of designs per object was $64 \times 20 = 1280$, the actual number of feasible designs generated was: 395 (curvy ball), 543 (gear), 149 (hourglass), 495 (spiky ball) and 489 (star). The hourglass geometry, being smoother than the `spikier' objects, typically produced simpler contact force distributions, causing SimTO to converge in fewer than 3 iterations on average. For the remaining objects, SimTO converged in approximately 6 to 9 iterations on average. An additional 245 infeasible designs - 10.6\% of the total - were automatically excluded from the analysis, either because they had disconnected geometries or lacked material at the fixed port of the design domain. 

In all optimization runs, the design domain from Fig.~\ref{fig:designdomain} was used as the initial simulated design. Dynamic grasping simulations spanned $t = 4s$, where grippers were first compressed by $d_c = 80\,\text{mm}$ during the first two seconds, before being lifted $d_l = 30\,\text{mm}$. Across all SimTO iterations, the initial gripper grasp pose $\mathcal{P}_g$ and object pose $\mathcal{P}_o$ remained constant. Maintaining fixed poses was not mandatory, and we hypothesize that strategically varying $\mathcal{P}_g$ and $\mathcal{P}_o$ would produce morphologies that are more robust to object-gripper pose variations.

\subsection{Analyzing the trade-off between geometric diversity and object lift time}

\begin{figure}[th!]
    \centering
    \includegraphics[width=0.50\textwidth]{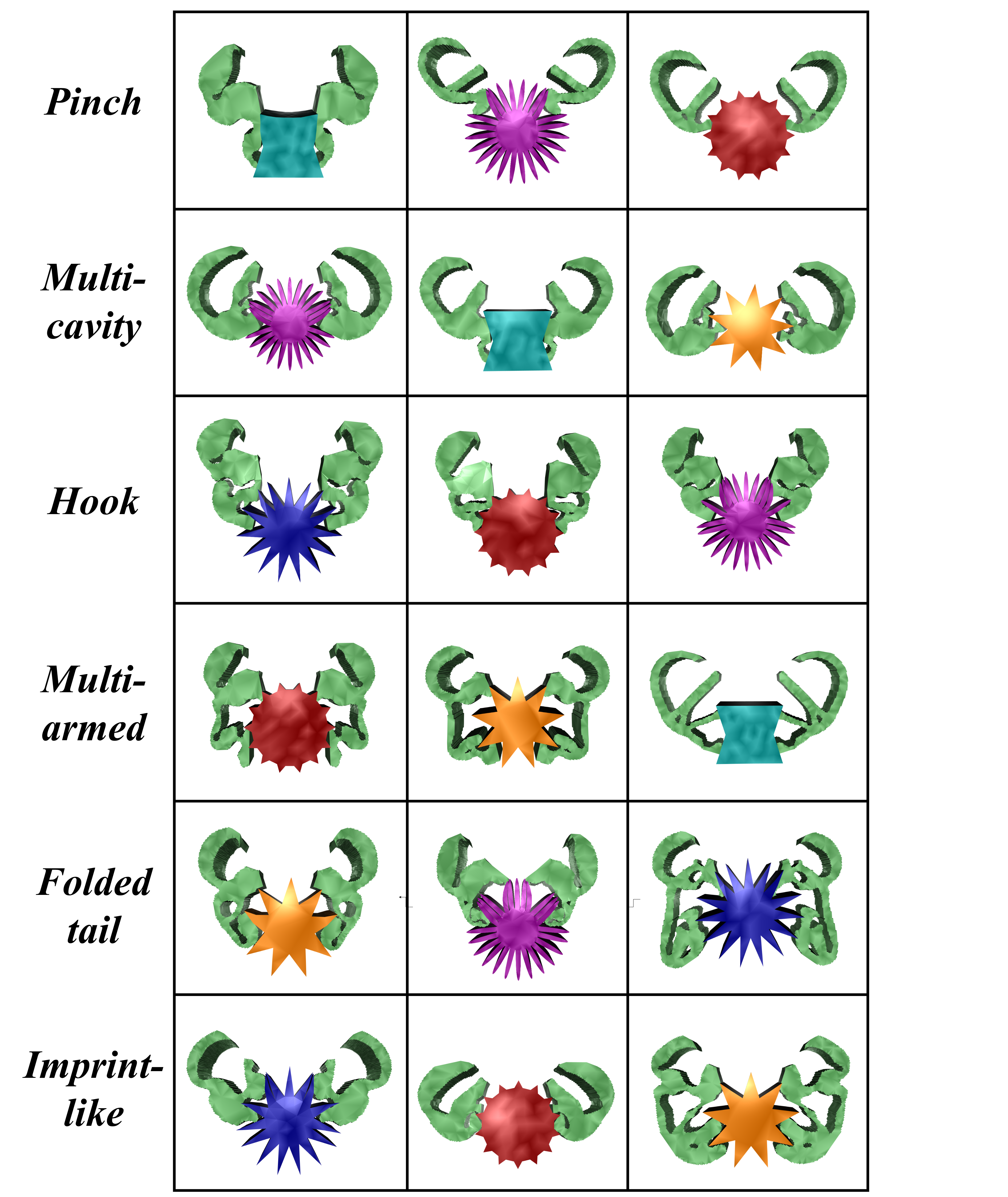}
    \caption{Six soft grasping strategies discovered by SimTO, with three examples shown per strategy:
    \textcolor{black}{(1) \textbf{Pinch}: opposing fingers apply large normal forces on objects;
    (2) \textbf{Multi-cavity}: multiple cavities capture object protrusions; 
    (3) \textbf{Hook}: a curved hook feature firmly encases object protrusions;
    (4) \textbf{Multi-armed}: multiple appendages secure the object at several sites; 
    (5) \textbf{Folded tail}: compliant tails fold inward to help carry the object; 
    (6) \textbf{Imprint-like}: gripper deforms to match complex object geometry. Supplementary videos illustrating these grasping strategies are available at: \url{https://kurtenkera.github.io/SimTO/}.}}
    \label{fig:diverse_grasping_modes}
\end{figure}

The relationship between \emph{geometric diversity} and \emph{object lift time} is illustrated in Fig.~\ref{fig:results}. For each feature-rich object, the geometric diversity of the $i$th design was defined as $D_i = \| \bm{\rho}_i - \bar{\bm{\rho}}\|$, where $\bar{\bm{\rho}} = \frac{1}{K}\sum_{i=1}^{K}\bm{\rho}_i$ represents the mean morphology of the $K$ designs produced for that object. Higher geometric diversity values correspond to designs that deviate significantly from the mean morphology. We used \emph{object lift time} as a proxy for functional performance, which was defined as the total duration that the object was simultaneously in contact with the gripper while not touching the ground. 

As shown in Fig.~\ref{fig:results}, designs located on the Pareto front typically used a high volume fraction of 0.30 or 0.35. Only the spiky ball had one Pareto-optimal design with $v_f = 0.25$. This trend was expected, as a larger material budget gives the optimizer the freedom to generate more exotic morphologies with greater geometric diversity. Furthermore, higher volume designs are capable of generating larger contact forces, and are expected to increase object lift times due to their increased grip strength. 

While high geometric diversity often signals novel features tailored for specialized grasping, it can also indicate the presence of excessive or redundant features. This is evident in the curvy ball, spiky ball and star results (Fig.~\ref{fig:results}), where the most geometrically diverse designs developed long ``dead-weight'' appendages that dragged along the ground without contributing to grasping. Conversely, while designs with the highest lift times often possessed less diverse features that were functionally more useful, in some cases these designs demonstrated overfitting to the specific gripper-object pose combination used during optimization. Interestingly, designs located slightly off the Pareto front with relatively high object lift times and geometric diversity often possessed the most useful morphological features for robust and secure grasping, which we explicitly demonstrate in the next section.

\subsection{Evaluating the specificity and generality of twenty designs}\label{design_eval}

\begin{figure*}[th!]
    \centering
    \includegraphics[width=0.8\textwidth]{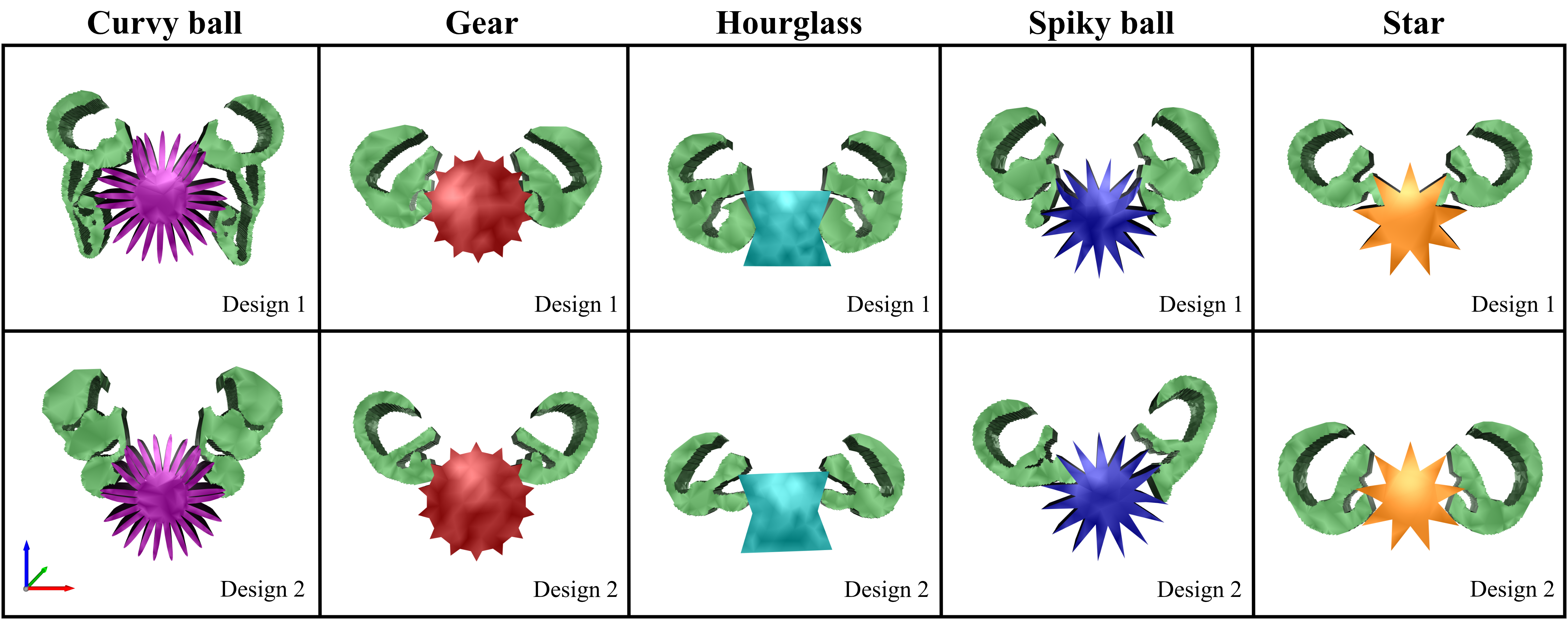}
    \caption{\emph{Ten soft designs}. All designs were optimized with $E_g = 11.51 \,\text{MPa}$ and evaluated with $E_g$ and $E_o$ also equal to $11.51\,\text{MPa}$ (see Table~\ref{tab1} for results). In this figure, each design is compressed by $d_c = 80\, \text{mm}$.}
    \label{fig:exp1designs}
\end{figure*}

\begin{figure*}[th!]
    \centering
    \includegraphics[width=0.8\textwidth]{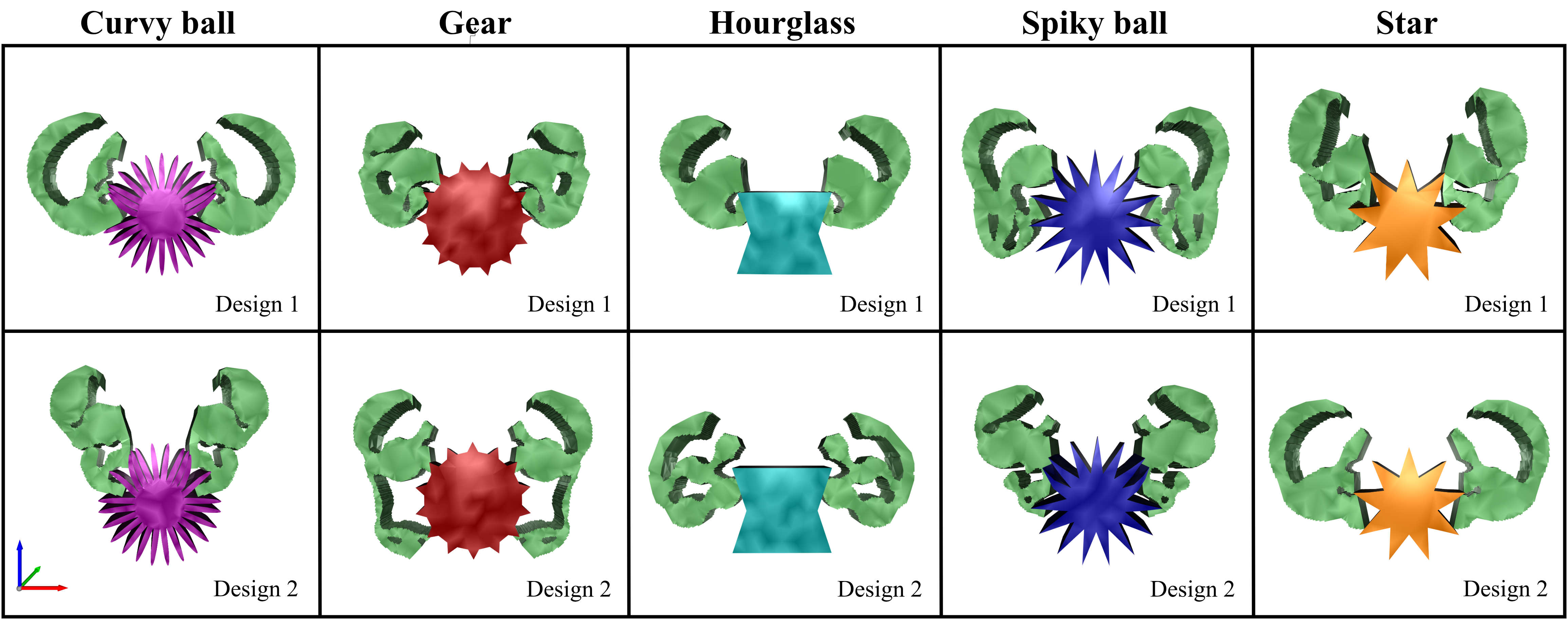}
    \caption{\emph{Ten softer designs}. All designs were evaluated with $E_g$ and $E_o$ set to $0.46\,\text{MPa}$ (see Table~\ref{tab2}). In this figure, each design is compressed by $d_c = 80\, \text{mm}$.}
    \label{fig:exp2designs}
\end{figure*}

From the results presented in Fig.~\ref{fig:results}, we manually selected twenty designs for further evaluation: ten \emph{soft} designs (Fig.~\ref{fig:exp1designs}) optimized with $E_g = 11.51 \, \text{MPa}$ and ten \emph{softer} designs (Fig.~\ref{fig:exp2designs}) optimized with $E_g \in \{0.46, 0.60, 1.39\} \, \text{MPa}$, with two designs per object in each group.

In-domain and out-of-domain grasping performance were evaluated for all designs, with $E_g = E_o = 11.51 \, \text{MPa}$ for all \emph{soft} designs, and $E_g = E_o = 0.46 \, \text{MPa}$ for all \emph{softer} designs. For in-domain testing, we performed 35 simulated grasps per design on its target object. Seven object poses were used: rotations of $5^\circ$ and $10^\circ$ about the $z$-axis; translations of $6\, \text{mm}$ and $12\, \text{mm}$ along the $x$-axis; translations of $8\, \text{mm}$ and $16\, \text{mm}$ along the $y$-axis, and the origin pose. Each pose was evaluated in five independent trials with distinct random seeds. For out-of-domain testing, each design was evaluated on four unseen objects, again with 35 grasps per object.


\begin{table*}[h]
\caption{\emph{Results for ten soft designs} (Fig.~\ref{fig:exp1designs}). Optimization parameter values are indicated by ``Opt.'' where applicable to distinguish them from evaluation values.} \label{tab1}
\begin{tabular}{lccccccccc}
\toprule%
\thead{\textbf{Object}} & \thead{\textbf{Design}} & 
\thead{\bm{$E_g$}} &
\thead{\bm{$E_o$}\\\textbf{(Opt.)}}  &
\thead{\bm{$v_f$}} &
\thead{\textcolor{black}{\bm{$N_f$}}} &
\thead{\textbf{SimTO}\\\textbf{Iters.}} & 
\thead{\textbf{In-domain}\\\textbf{Object Stress (Pa)}} & 
\thead{\textbf{In-domain}\\\textbf{Success}} &
\thead{\textbf{Out-of-domain}\\\textbf{Success}} \\
\midrule
\small{\textbf{Curvy}} & 1 & 11.51 & 0.46 & 0.25 & \textcolor{black}{36} & 4  & $(7.42 \pm 1.61) \times 10^{5}$ & 80.00\% & 89.29\%  \\
\small{\textbf{ball}}& 2 & 11.51 & 0.46 & 0.35 & \textcolor{black}{61} & 2  & $(9.77 \pm 2.53) \times 10^{5}$ & 74.29\% &  71.43\%  \\
\cmidrule(lr){1-10}
\small{\textbf{Gear}} & 1 & 11.51 & 1.39 & 0.35 & \textcolor{black}{21} & 5  & $(2.03 \pm 0.59) \times 10^{5}$ & 91.43\% & 77.86\% \\
& 2 & 11.51 & 11.51 & 0.20 & \textcolor{black}{12} &  9  & $(1.20 \pm 0.22) \times 10^{5}$ & 88.57\% & 82.14\% \\
\cmidrule(lr){1-10}
\small{\textbf{Hour-}} & 1 & 11.51 & 0.46 & 0.35 & \textcolor{black}{42}  & 5  & $(1.07 \pm 0.25) \times 10^{5}$ & 94.29\% & 76.43\% \\
\small{\textbf{glass}}& 2 & 11.51 & 1.39 & 0.20 & \textcolor{black}{22} & 2  & $(1.34 \pm 0.41) \times 10^{5}$ & 85.71\% & 89.29\% \\
\cmidrule(lr){1-10}
\small{\textbf{Spiky}} & 1 & 11.51 & 0.46 & 0.30 & \textcolor{black}{35} & 6  & $(7.80 \pm 4.09) \times 10^{5}$ & 74.29\% & 82.14\%  \\
\small{\textbf{ball}}& 2 & 11.51 & 1.39 & 0.20 & \textcolor{black}{24} & 4  & $(4.36 \pm 1.49) \times 10^{5}$ & 71.43\% & 89.29\% \\
\cmidrule(lr){1-10}
\small{\textbf{Star}} & 1 & 11.51 & 1.39 & 0.20 & \textcolor{black}{26} & 1  & $(1.88 \pm 0.17) \times 10^{5}$ & 85.71\% & 84.29\%  \\
& 2 & 11.51 & 11.51 & 0.30 & \textcolor{black}{21} & 2  & $(4.25 \pm 1.09) \times 10^{5}$ & 85.71\% &  87.86\%  \\
\botrule
\end{tabular}
\end{table*}


During evaluation, each simulated grasp lasted $t=8s$, with the first $4s$ used to compress the gripper by $d_c = 80\,\text{mm}$, and the remaining time used to lift the gripper by $40\,\text{mm}$. A grasp was considered successful if, at the end of the simulation, the object remained in contact with the gripper while not touching the ground.  During in-domain tests, peak von Mises stresses were recorded for the objects being grasped. Tables~\ref{tab1} and~\ref{tab2} summarize the results. All in-domain results are averages over 35 independent grasps, and out-of-domain success rates are based on 140 grasps across four unseen objects. 

\subsubsection{Grasping performance of soft designs}

All soft designs achieved high in-domain and out-of-domain grasp success rates (Table~\ref{tab1}), indicating robustness to object pose variations and an ability to generalize to unseen objects. While most designs required multiple SimTO iterations to develop their features, some high-performing designs emerged in as few as 1-2 iterations (e.g., the second hourglass and both star designs). \textcolor{black}{Notably, the number of contact-based loads used to optimize these designs, $N_f$, reached as high as 61 for the second curvy ball design. This is in stark contrast to conventional `dummy load' approaches for soft gripper design, which typically use only 1-5 manually specified loads.}

\subsubsection{Grasping performance of softer designs}


\begin{table*}[h]
\caption{\emph{Results for ten softer designs}. Optimization parameter values are indicated by ``Opt.'' where applicable to distinguish them from evaluation values.} \label{tab2}
\begin{tabular}{lccccccccccc}
\toprule%
\thead{\textbf{Object}} & \thead{\textbf{Design}} & 
\thead{\bm{$E_g$}\\\textbf{(Opt.)}}  &
\thead{\bm{$E_o$}\\\textbf{(Opt.)}}  &
\thead{\bm{$v_f$}} &
\thead{\textcolor{black}{\bm{$N_f$}}} &
\thead{\textbf{SimTO}\\\textbf{Iters.}} & 
\thead{\textbf{In-domain}\\\textbf{Object Stress (Pa)}} & 
\thead{\textbf{In-domain}\\\textbf{Success}} &
\thead{\textbf{Out-of-domain}\\\textbf{Success}} \\
\midrule

\small{\textbf{Curvy}}  & 1 & 0.60 & 11.51 & 0.35 & \textcolor{black}{13} &  7  & $(3.37 \pm 0.63) \times 10^{4}$ & 71.43\% & 75.00\% \\
& 2 & 0.60 & 0.60 & 0.35 & \textcolor{black}{42} &  4 & $(5.71 \pm 1.44) \times 10^{4}$ & 71.43\% & 70.71\% \\
\cmidrule(lr){1-10}
\small{\textbf{Gear}} & 1 & 0.60 & 0.46 & 0.35 & \textcolor{black}{33} &  19  & $(1.16 \pm 0.21) \times 10^{4}$ & 71.43\% & 78.57\% \\
& 2 & 1.39 & 0.60 & 0.35 & \textcolor{black}{19} & 1  & $(1.31 \pm 0.40) \times 10^{4}$ & 71.43\% & 60.71\% \\
\cmidrule(lr){1-10}
\small{\textbf{Hour-}} & 1 & 1.39 & 0.60 & 0.30 &\textcolor{black}{11} &  2  & $(1.04 \pm 0.27) \times 10^{4}$ & 71.43\% & 71.43\% \\
\small{\textbf{glass}} & 2 & 0.60 & 11.51 & 0.25 &\textcolor{black}{56} & 3  & $(3.40 \pm 0.32) \times 10^{3}$ & 71.43\% & 60.00\% \\
\cmidrule(lr){1-10}
\small{\textbf{Spiky}} & 1 & 0.46 & 0.46 & 0.35 & \textcolor{black}{21}& 3  & $(2.60 \pm 0.69) \times 10^{4}$ & 71.43\% & 67.86\% \\
\small{\textbf{ball}}& 2 & 1.39 & 0.46 & 0.35 & \textcolor{black}{29}& 10  & $(4.28 \pm 1.83) \times 10^{4}$ & 57.14\% & 71.43\% \\
\cmidrule(lr){1-10}
\small{\textbf{Star}} & 1 & 0.46 & 0.46 & 0.35 & \textcolor{black}{34} & 11  & $(4.88 \pm 1.56) \times 10^{4}$ & 71.43\% & 46.43\% \\
& 2 & 0.46 & 1.39 & 0.35  & \textcolor{black}{6} & 13  & $(2.07 \pm 0.48) \times 10^{4}$ & 71.43\% & 60.71\% \\
\botrule
\end{tabular}
\end{table*}

The softer designs achieved slightly lower success rates (Table~\ref{tab2}) than the stiffer designs. This trend was expected, as reduced material stiffness inherently decreases overall grip strength. Despite this, most grippers maintained strong robustness to object pose variations and generalization to unseen objects. The highest out-of-domain success rates were associated with \textbf{multi-armed} (first hourglass and gear designs), \textbf{multi-cavity} (first curvy ball design), and \textbf{hook}-based (e.g., second spiky ball design) grasping strategies (Fig.~\ref{fig:diverse_grasping_modes}). Two multi-armed designs with strong in-domain and out-of-domain success rates also emerged in as few as 1-2 SimTO iterations (the second gear and first hourglass designs). This suggests that the use of rich contact-based loads alone can generate high-performing grippers, and that multiple SimTO iterations are not always required.

\section{Experimental Validation}\label{expval}

\begin{figure}[th!]
    \centering
    \includegraphics[width=0.41\textwidth]{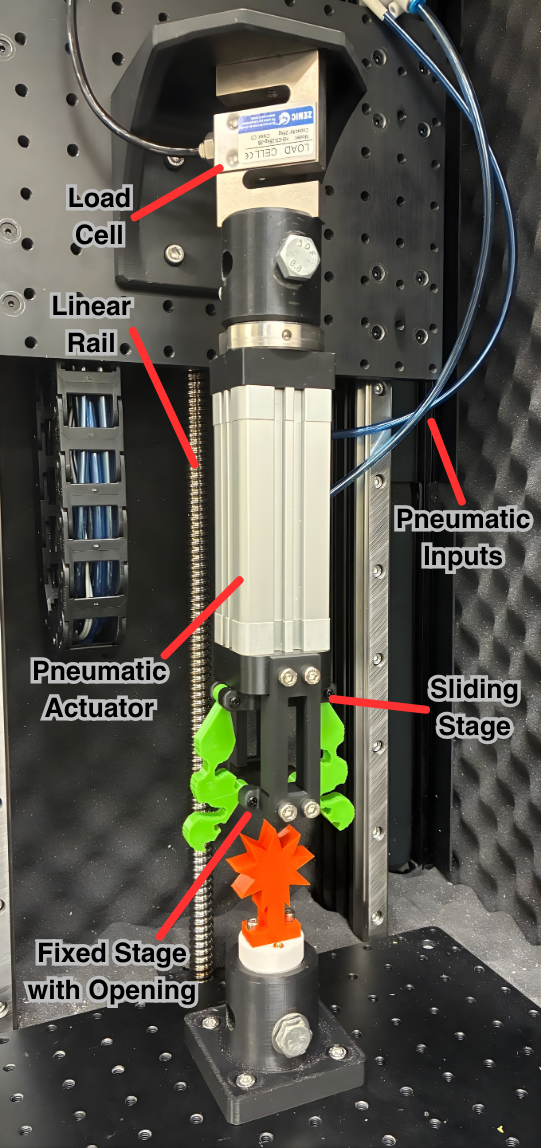}
    \caption{\textcolor{black}{Experimental setup for pull-off force testing. Our soft gripper was actuated by a pneumatic cylinder and positioned using a motorized linear rail.}}
    \label{fig:apparatus}
\end{figure}

We developed a physical soft robotic gripper (Fig.~\ref{fig:apparatus}) based on the design scheme proposed by \citet{liu2018_TObenchmark_v2}. To evaluate the specialization of our designs, we 3D-printed all ten designs from Fig.~\ref{fig:exp1designs} and measured their peak grasp forces on their corresponding target objects. We also fabricated the \emph{generalist} design by \citet{liu2018_TObenchmark_v2} (Fig.~\ref{fig:real_grasping_comparison}) and measured its peak grasp force across all five objects for comparison. 

This comparison is informative, as the generalist method uses conventional `dummy loads' and a fixed contact surface to approximate a range of objects and grasping conditions, whereas SimTO uses object-specific loads and allows contact surface geometries to evolve during optimization.

In these experiments, all soft fingers and objects were 3D-printed with 100\% infill using a Prusa FDM printer and NinjaTek thermoplastic polyurethane (TPU) filament with a Young's modulus of $12\,\text{MPa}$. To measure peak grasp forces, objects were rigidly fixed to our testing platform (Fig.~\ref{fig:apparatus}), and pull-off tests were performed by first (i) positioning the gripper above the object, (ii) closing the gripper, and then (iii) slowly lifting the gripper while recording the time-varying force in the vertical direction using a load cell. Five trials were conducted per design, and the average peak grasp force is reported in Table~\ref{tab3}. 


\begin{table}[t]
\caption{\textcolor{black}{\emph{Pull-off test results}. Reported values are mean $\pm$ standard deviation over five trials per design. ``Ref.'' denotes the generalist design of \citet{liu2018_TObenchmark_v2}.}} \label{tab3}
\begin{tabular}{lcc}
\toprule%
\thead{\textbf{Object}} & \thead{\textbf{Design}} & 
\thead{\textbf{Peak Grasp}\\\textbf{Force (N)}} \\
\midrule

\small{\textbf{Curvy ball}}  & 1 & $6.51 \pm 0.91$ \\
& \textbf{2} & $\bf{37.17 \pm 3.19}$ \\
& Ref. & $4.18 \pm 0.27$ \\
\cmidrule(lr){1-3}
\small{\textbf{Gear}} & 1 & $26.56 \pm 1.43$ \\
& \textbf{2} & $\bf{46.20\pm 2.81}$ \\
& Ref. & $7.04 \pm 0.50$ \\
\cmidrule(lr){1-3}
\small{\textbf{Hourglass}} & 1 & $20.69 \pm 0.71$ \\
& \textbf{2} & $\bf{28.29 \pm 3.08}$ \\
& Ref. & $20.94 \pm 2.31$ \\
\cmidrule(lr){1-3}
\small{\textbf{Spiky ball}} & \textbf{1} & $\bf{30.72 \pm 3.42}$ \\
& 2 & $16.50 \pm 0.13$ \\
& Ref. & $5.04 \pm 0.24$ \\
\cmidrule(lr){1-3}
\small{\textbf{Star}} & 1 & $21.49 \pm 0.61$ \\
& \textbf{2} & $\bf{76.32 \pm 3.62}$ \\
& Ref. & $7.89 \pm 0.14$ \\
\botrule
\end{tabular}
\end{table}

\subsection{Generating superior contact with specialized grippers}

\begin{figure*}[th!]
    \centering
    \includegraphics[width=0.95 \textwidth]{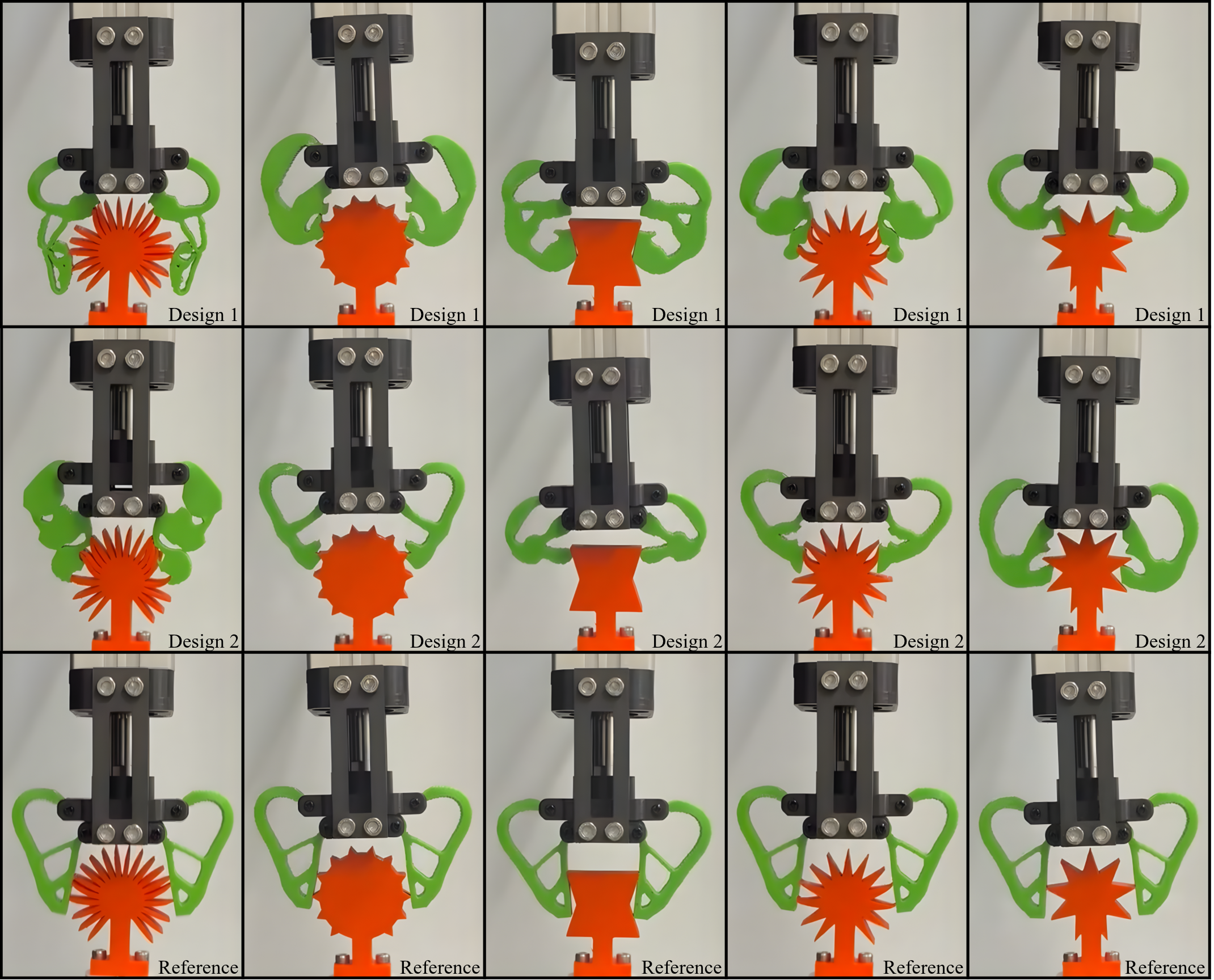}
    \caption{\textcolor{black}{Representative snapshots of each design during pull-off tests illustrating the contact configurations formed between the gripper and object during grasping. The generalist reference design consistently generated lower contact area with feature-rich objects due to its flat contact surface.}}
    \label{fig:real_grasping_comparison}
\end{figure*}

Table~\ref{tab3} shows that the evaluated specialized designs achieved significantly higher grasp forces than the generalist design. This difference was primarily due to the generalist design's inability to establish a large contact area with feature-rich objects during grasping (Fig.~\ref{fig:real_grasping_comparison}). Its long, flat contact surface was unable to conform to object grooves during grasping, and instead it relied on applying large normal forces on object protrusions (i.e. `pinch' grasping). 

In contrast, our specialized designs achieved a large contact area during grasping due to their object-specific contact surfaces. For example, the second gear design used its thumb-like features to firmly engage the gear teeth during pull-off tests (Fig.~\ref{fig:real_grasping_comparison}), while the second curvy ball and first spiky ball designs employed hook-based grasping strategies to latch onto several object protrusions. The second star design also reached beneath the outer lobes of the star, producing the highest measured grasp force of $76.32 \,\text{N}$.

\section{Discussion and Conclusions}\label{sec5}

We have presented SimTO, a two-stage, \textbf{sim}ulation-driven \textbf{T}opology \textbf{O}ptimization framework that removes the need for pre-defined load cases by automatically extracting them from dynamic physics simulations. By repeatedly alternating between dynamic, simulation-based load extraction and classical topology optimization, we show that SimTO generates diverse gripper designs and distinct grasping strategies.

A key feature of SimTO is that the gripper contact surface is not prescribed \emph{a priori}, but instead emerges during optimization. This enables the discovery of designs with object-specific contact surfaces. Experimental results show that such designs can achieve substantially higher grasp forces than generalist designs produced by conventional topology optimization methods.

One limitation of our current approach is its susceptibility to producing disconnected geometries, a known issue in density-based topology optimization \citep{Cool2025}. Incorporating connectivity constraints \citep{Cool2025} is a promising direction to address this. In addition, design selection in this work was performed manually (Sec.~\ref{design_eval}). However, soft robotic grippers deform and interact with objects in non-intuitive ways, meaning manual selection risks overlooking high-performing designs that use unconventional or unintuitive grasping modes. Developing automatic design-selection criteria is therefore another important direction for future work.

\textcolor{black}{Several further extensions are also promising. First, SimTO could be used to independently optimize multiple unique fingers within a single soft gripper, enabling asymmetric grippers with novel grasping strategies. Secondly, 3D topology optimization could be used to generate new designs with features that are unattainable with a 2D formulation. Third, the gripper grasp pose and object pose could be strategically varied during design optimization to create designs that are more robust to object and gripper positioning uncertainty.} Finally, while SimTO achieves strong results using linear analysis during topology optimization, nonlinear finite element modelling would give a higher fidelity optimization result at the cost of increased computational complexity.

Overall, SimTO provides a framework for generating soft grippers specialized to feature-rich objects, and our results highlight the advantages of specialization over generic design. Such designs are particularly relevant for industrial soft grasping applications involving constrained object sets, such as fruit harvesting, manufacturing, and warehousing, where task-specific gripper design can yield substantial performance gains.

\backmatter

\color{black}
\section*{Declarations}

\bmhead{Author contributions}
All authors contributed to the study conception and design. Material preparation, data collection and analysis were performed by Kurt Enkera. The first draft of the manuscript was written by Kurt Enkera and all authors commented on previous versions of the manuscript. All authors read and approved the final manuscript.

\bmhead{Conflict of interest} 
We have no conflict of interest to report.

\bmhead{Data availability} 
All designs presented in this paper are available at: \url{https://github.com/kurtenkera/SimTO-Dataset}.

\bmhead{Replication of results} 
The results presented in this paper can be replicated by implementing the methods outlined in Sections~\ref{method} and~\ref{results}.

\bmhead{Funding} 
Not applicable.

\bmhead{Ethics approval and consent to participate} 
Not applicable.

\bibliography{sn-bibliography}
\end{document}